\documentclass[runningheads]{llncs}

\usepackage[mobile]{eccv}

\usepackage{eccvabbrv}

\usepackage{graphicx}
\usepackage{booktabs}

\usepackage[accsupp]{axessibility}  

\usepackage{color}
\usepackage{colortbl}

\usepackage{amsmath}
\usepackage{amssymb}
\usepackage{makecell}
\usepackage{array}
\usepackage{wrapfig}
\usepackage{multirow}
\usepackage{bbm}
\usepackage{float}
\usepackage{graphicx}
\usepackage{perpage}
\usepackage{enumitem}
\usepackage{algpseudocode}
\usepackage[ruled,vlined]{algorithm2e}
\usepackage{extarrows}
\usepackage{amssymb}
\usepackage{booktabs}
\usepackage{marvosym}
\usepackage{pifont}
\usepackage{gensymb}

\usepackage{listings}

\definecolor{LightCyan}{gray}{0.9}
\newcolumntype{a}{>{\columncolor{LightCyan}}c}

\usepackage[pagebackref,breaklinks,colorlinks,citecolor=eccvblue]{hyperref}

\usepackage{orcidlink}

\makeatletter\renewcommand\paragraph{\@startsection{paragraph}{4}{\z@}
  {.5em \@plus1ex \@minus.2ex}{-.5em}{\normalfont\normalsize\bfseries}}\makeatother

\begin{document}

\title{ULU: A Unified Activation Function
} 

\titlerunning{ULU}

\author{Simin Huo\inst{1}\orcidlink{0009-0008-7517-7915}}

\authorrunning{Simin Huo.~Author.}

\institute{Shanghai Jiao Tong University\\
\email{sameenhuo@sjtu.edu.cn}
}

\maketitle

\begin{abstract}

We propose \textbf{ULU}, a novel non-monotonic, piecewise activation function defined as $\{f(x;\alpha_1),x<0; f(x;\alpha_2),x>=0 \}$, where $f(x;\alpha)=0.5x(tanh(\alpha x)+1),\alpha >0$. ULU treats positive and negative inputs differently. Extensive experiments demonstrate ULU significantly outperforms ReLU and Mish across image classification and object detection tasks. Its variant Adaptive ULU (\textbf{AULU}) is expressed as $\{f(x;\beta_1^2),x<0; f(x;\beta_2^2),x>=0 \}$, where $\beta_1$ and $\beta_2$ are learnable parameters, enabling it to adapt its response separately for positive and negative inputs. Additionally, we introduce the LIB (Like Inductive Bias) metric from AULU to quantitatively measure the inductive bias of the model.

\end{abstract}

\section{Introduction}
\label{sec:intro}

Activation functions are elemental in imparting non-linearity to neural networks. Within a network architecture, the linear transformed inputs are fed into activation functions to generate non-linear outputs. These non-linear element-wise functions profoundly influence model performance. Hence, selecting appropriate activation functions to enable efficacious training and optimization has remained an engaging research area. In early neural networks, the \textit{sigmoid} ~\cite{Cybenko1989ApproximationBS}  and  \textit{tanh} ~\cite{LeCun1998EfficientLB} activation functions were widely used ~\cite{LeCun1989BackpropagationAT},~\cite{Hochreiter1997LongSM},~\cite{Graves2013SpeechRW}. However, they suffered from limitations like being upper bounded and causing vanishing gradients ~\cite{hochreiter1998vanishing}, ~\cite{glorot2010understanding}, which constrained model expressiveness.

To overcome some of the disadvantages of \textit{sigmoid} and \textit{tanh}, the Rectified Linear Unit,  ReLU~\cite{Nair2010RectifiedLU} was proposed. ReLU was simpler, easier to optimize, and demonstrated better generalization and faster convergence. This led to its widespread adoption in neural networks ~\cite{krizhevsky2012imagenet}, ~\cite{he2016deep}. However, ReLU also exhibited shortcomings, most notably the "dying ReLU" problem ~\cite{glorot2010understanding}. By collapsing all negative inputs to zero, ReLU could cause gradient information loss and stall model training. Moreover, the non-differentiability at $x=0$ could lead to optimization difficulties.

To address these issues, many refinements to ReLU were proposed

\textbf{Leaky ReLU}~\cite{Maas2013RectifierNI} incorporated a small negative slope to mitigate the "dying ReLU" problem. It has been used in many applications with promising performance. One major problem associated with Leaky ReLU is the finding of the right slope in linear function for negative inputs. Different slopes might be suited for different problems and different networks. \textbf{PReLU}~\cite{He2015DelvingDI} considers the slope for negative input as a trainable parameter. However, it can lead to overfitting easily which is the downside of PReLU. \textbf{ELU}~\cite{Clevert2015FastAA} exhibits all the benefits of the ReLU function. It is differentiable, saturates for large negative inputs and reduces the bias shift. The negative saturation regime of ELU adds some robustness to noise as compared to the Leaky ReLU and Parametric ReLU. \textbf{SELU}~\cite{Klambauer2017SelfNormalizingNN} extend ELU by using a scaling hyperparameter to make the slope larger than one for positive inputs. Basically, the SELU induces self-normalization to automatically converge towards zero mean and unit variance. \textbf{GELU} ~\cite{Hendrycks2016GaussianELU} considers nonlinearity as the stochastic regularization driven transformation. The complexity of GELU increases due to use of probabilistic nature. \textbf{Swish} ~\cite{Ramachandran2017SearchingFA} introduce self-gating to balance linear and non-linear behavior. The smaller and higher values of hyperparameter lead towards the linear and ReLU functions, respectively. Thus, it can control the amount of non-linearity based on the dataset and network complexity. \textbf{SiLU} ~\cite{elfwing2018sigmoid} make the output of the sigmoid function multiply with its input in sigmoid-weighted linear unit. \textbf{Mish} ~\cite{Misra2019MishAS} combine the softplus function with Tanh function together. It is non-monotonic and smooth. However, the increased complexity in Mish due to the multiple functions can be a limitation for the deep networks.

In this work, we propose novel non-monotonic activation functions Unified Linear Unit (\textbf{ULU}) and Adaptive ULU (\textbf{AULU}), inspired by the shape of Mish. The biggest feature of ULU is that it treats positive and negative inputs differently, whereas AULU processes them in an adaptive manner. In addition, we introduce the LIB (Like Inductive Bias) metric from AULU to quantitatively measure the inductive bias of the model. Through extensive experiments, we demonstrate that our proposed AULU surpasses ReLU and GELU across various tasks including image classification and object detection.
\section{Related Work}
\label{sec:related}

Activation functions play a crucial role in deep neural networks, introducing non-linearity and enabling the modeling of complex data representations. The Rectified Linear Unit (ReLU) has been widely adopted due to its simplicity and effectiveness in alleviating the vanishing gradient problem. However, ReLU~\cite{Nair2010RectifiedLU} and its variants are not without limitations, prompting researchers to explore alternative activation functions.

One line of research focuses on addressing the non-utilization of negative values in ReLU. The Leaky ReLU (LReLU)~\cite{Maas2013RectifierNI} introduces a small negative slope to mitigate the "dying ReLU" problem, but determining the optimal slope remains a challenge. Parametric ReLU (PReLU) addresses this by treating the negative slope as a trainable parameter, although it can lead to overfitting. Other variants, such as Randomized ReLU (RReLU)\cite{rrelu}, Concatenated ReLU (CReLU)\cite{crelu}, and Parametric Tan Hyperbolic Linear Unit (P-TELU)\cite{ptelu}, attempt to capture useful information from negative inputs.

Another research direction aims to enhance the limited non-linearity of ReLU. S-shaped ReLU (SReLU) and Multi-bin Trainable Linear Unit (MTLU)\cite{mtlu} increase non-linearity by combining multiple linear functions. Elastic ReLU (EReLU)\cite{erelu} controls non-linearity by randomly drawing slopes during training. Rectified Linear Tanh (ReLTanh)\cite{reltanh}combines ReLU with Tanh to overcome vanishing gradients.

The unbounded output of ReLU and its variants has also been a concern, as it may lead to training instability, particularly in embedded systems. Bounded ReLU (BReLU) \cite{BReLU} addresses this issue by introducing an upper bound, improving training stability.

Exponential activation functions, such as Exponential Linear Unit (ELU) \cite{elu}, Scaled ELU (SELU)\cite{selu}, Parametric ELU (PELU)\cite{pelu}, and their variants, tackle the gradient diminishing problem of ReLU. These functions exhibit desirable properties like differentiability, negative saturation, and reduced bias shift, providing robustness to noise and improved optimization.

Despite the numerous ReLU variants proposed, there is still a need for activation functions that can effectively capture the intricate patterns present in complex data while addressing the limitations of existing functions. The development of novel activation functions that balance non-linearity, adaptability, and computational efficiency remains an active area of research in deep learning.
\section{Method}
\label{sec:method}

\subsection{Motivation}
\textit{ULU} proposed is inspired by the shape of the function \textit{Mish}. Observe the mathematical expression of the \textit{Mish} :

\begin{equation}
Mish(x) =  x \cdot tanh(\ln (1 + {e^x}))
\label{eq1}
\end{equation}

It can be observed that when $x$ approaches negative infinity, Mish value converges to zero. 
\begin{equation}
\mathop {\lim }\limits_{x \to  - \infty } Mish(x) = 0
\label{eq2}
\end{equation}

As $x$ tends towards positive infinity, Mish approximates the following expression.
\begin{equation}
\mathop {\lim }\limits_{x \to  + \infty } Mish(x) = x \cdot tanh(x)
\label{eq3}
\end{equation}

\begin{figure}[tb]
    \centering
    \includegraphics[width = 0.23\textwidth,height = 0.23\textwidth]{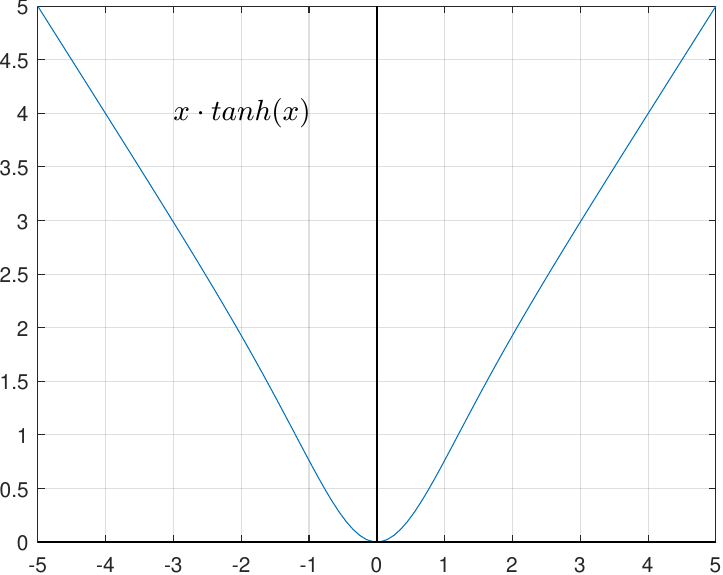}
    \hspace{0cm}
    \includegraphics[width = 0.23\textwidth,height = 0.23\textwidth]{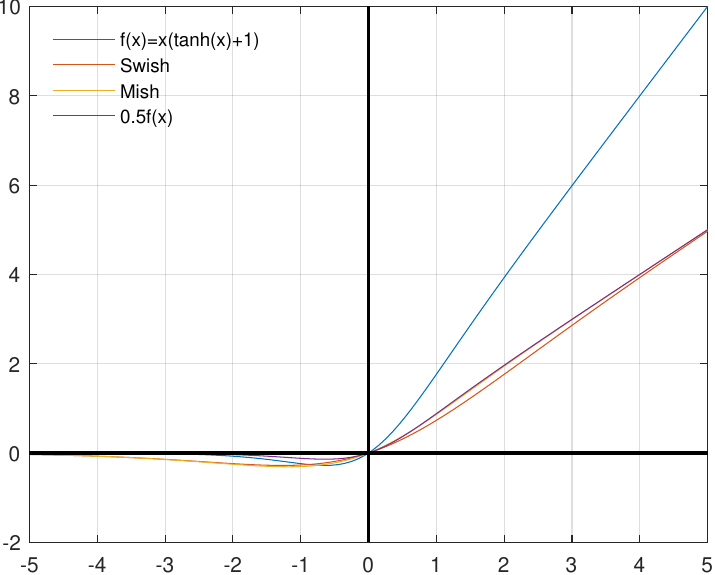}
    \caption{ The graph of the ~\cref{eq3} (Left) and ~\cref{eq4}, ~\cref{eq8}, Swish and Mish (Right)}%
    \label{fig:curves}
\end{figure}

The graph of the~\cref{eq3} reveals it is an even function. To make its output resemble \textit{ReLU}, \textit{Swish} and \textit{Mish} for negative inputs, a negative number needs to be added to the output when $x < 0$, which means
\begin{equation}
\mathop {\lim }\limits_{x \to  - \infty } x\cdot tanh(x) + C = -x + C = 0
\label{eq3_1}
\end{equation}
Therefore, we opt to add $x$ itself. This leads to the formulation:
\begin{equation}
f(x) = x \cdot tanh(x) + x = x(tanh(x) + 1)
\label{eq4}
\end{equation}
Its $1^{st}$ and $2^{nd}$ derivative of~\cref{eq4} are as follows:
\begin{align}
 f'(x) & = tanh(x) - x \cdot (tan{h^2}(x) - 1) + 1 \notag\\
 & = tanh(x) +x \cdot sec{h^2}(x) + 1
\label{eq5}
\end{align}
\begin{align}
f''(x) &= - tan{h^2}(x) + x \cdot tanh(x)(tan{h^2}(x) - 1) + 1 \notag\\
& = 2sec{h^2}(x)(1-tanh(x))
\label{eq6}
\end{align}

In order to constrain the integral of $f''(x)$  over the negative and positive infinity intervals to 1, analogous to a probability density function integrating to 1, we need to multiply the integral result of $f''(x)$ by a constant $s$. Through the following derivation, we obtain that the constant $s = 0.5$:
\begin{align}
s\int_{ - \infty }^{ + \infty } {f''(x)dx}  &= s(f'( + \infty ) - f'( - \infty )) \notag \\
&= s(2 - 0) = 1
\label{eq7}
\end{align}
Therefore, Formula~\cref{eq4} is revised as follows:
\begin{equation}
s\cdot f(x) =sx(tanh(x) + 1) = 0.5x(tanh(x) + 1)
\label{eq8}
\end{equation}

The formula~\cref{eq8} possesses excellent properties similar to ReLU (\textit{i.e}, $f'(+\infty)=1$). We may wish to further consider a more general form of~\cref{eq8}. The constructed function is as follows:
\begin{equation}
g(x) = 0.5x(tanh(\alpha x) + 1),  \alpha \neq 0
\label{eq9}
\end{equation}
 Then the first derivatives of the function $g(x)$
\begin{equation}
g'(x) = 0.5(\alpha \cdot sec{h^2}(\alpha x) + tanh( \alpha x) + 1)
\label{eq10}
\end{equation}
When $\alpha >0$
\begin{align}
g( + \infty )=x,  g( - \infty ) = 0 \notag \\
g'( + \infty )=1,  g'( - \infty ) = 0 
\label{eq:agreat0}
\end{align}
When $\alpha <0$
\begin{align}
g( + \infty ) =0,  g( - \infty ) = -x  \notag \\
g'( + \infty ) =0,  g'( - \infty ) = -1 
\label{eq:aless0}
\end{align}

We discard~\cref{eq:aless0} because the activation function $g(x)$ is supposed to have properties similar to the ReLU function, \textit{i.e}, $g'(+ \infty) = 1, g'(- \infty)=0$. So we update $g(x)$
\begin{equation}
    g(x) =  0.5x(tanh(\alpha x) + 1), \,\, \alpha > 0
    \label{eq:gx}
\end{equation}

The~\cref{eq:agreat0} shows that the parameter $\alpha$ value has no effect on $g'(+ \infty) = 1, g'(- \infty)=0$, as long as $\alpha > 0$. Therefore, in order to make the activation function exhibit \textbf{different behaviors over different input regions}, we can construct a piecewise function as follows:
\begin{equation}
h(x)= \begin{cases}
    0.5x(tanh(\alpha_1 x) + 1), & \text{if } x < a \\
    0.5x(tanh(\alpha_2 x) + 1), & \text{if } x \geq a
\end{cases}   
\label{eq:g}
\end{equation} 

We have to make sure integral of $h''(x)$ over the negative and positive infinity intervals to 1

\begin{align}
\int_{-\infty}^{+\infty} h''(x) \, dx &= \left. h'(x) \right|_{-\infty}^a + \left. h'(x) \right|_a^{+\infty} \notag \\
&= 0.5 \bigg( \tanh(\alpha_1 a) + \alpha_1 a \cdot sec{h^2}(\alpha_1 a) \notag \\
&\quad - \tanh(\alpha_2 a) - \alpha_2 a \cdot sec{h^2}(\alpha_2 a) \bigg) + 1 = 1 
\label{eq:hx}
\end{align}

which means $\Rightarrow a = 0$

So when $a = 0$, we have the following activation function named \textbf{ULU}, which means \textbf{U}nited \textbf{L}inear \textbf{U}nit:

\begin{equation}
ULU= \begin{cases}
    0.5x(tanh(\alpha_1 x) + 1), & \text{if } x < 0 \\
    0.5x(tanh(\alpha_2 x) + 1), & \text{if } x \geq 0
\end{cases}   
\label{eq:ULU}
\end{equation}

 $\alpha_1 , \alpha_2 > 0$.
 
However, while $\alpha_1$ and $\alpha_2$ provide ULU with flexibility according to the tasks and models, they also decrease its reliability, as manually selecting suitable $\alpha_1$ and $\alpha_2$ values before training is difficult. Therefore, we propose an adaptive ULU (\textbf{AULU}), with the following expression:

\begin{equation}
AULU= \begin{cases}
    0.5x(tanh(\beta_1^2 x) + 1), & \text{if } x < 0 \\
    0.5x(tanh(\beta_2^2 x) + 1), & \text{if } x \geq 0
\end{cases}   
\label{eq:AULU}
\end{equation}

Where $\beta_1,\beta_2$ are learnable parameters.  whose squares ensure the coefficients preceding $x$ stay positive.

Note that
\begin{equation}
    tanh (x) = 2 \cdot \sigma (2x) - 1
\end{equation}

So ~\cref{eq:ULU} and ~\cref{eq:AULU} can be rewritten as:

\begin{equation}
ULU= \begin{cases}
    x\cdot \sigma(2 \alpha_1 x), & \text{if } x < 0 \\
    x\cdot \sigma(2 \alpha_2 x), & \text{if } x \geq 0
\end{cases}   
\label{eq:ULU2}
\end{equation}

\begin{equation}
AULU= \begin{cases}
    x\cdot \sigma(2 \beta_1^2 x), & \text{if } x < 0 \\
    x\cdot \sigma(2 \beta_2^2 x), & \text{if } x \geq 0
\end{cases}   
\label{eq:AULU2}
\end{equation}

Because 2 can be absorbed by the hyperparameter $ \alpha$ and  $\beta^2$

So we have more concise expressions
\begin{equation}
ULU= \begin{cases}
    x\cdot \sigma(\alpha_1 x), & \text{if } x < 0 \\
    x\cdot \sigma(\alpha_2 x), & \text{if } x \geq 0
\end{cases}   
\label{eq:ULU3}
\end{equation}

\begin{equation}
AULU= \begin{cases}
    x\cdot \sigma(\beta_1^2 x), & \text{if } x < 0 \\
    x\cdot \sigma(\beta_2^2 x), & \text{if } x \geq 0
\end{cases}   
\label{eq:AULU3}
\end{equation}

The expressions are very similar to Swish
\begin{equation}
Swish= x\cdot \sigma(\gamma x) 
\label{eq:swish}
\end{equation}

where $\gamma$ can be a constant or trainable parameter.

The primary distinctions between the Swish, ULU, and AULU activation functions lie in two key areas: their function structure (whether they are single or piecewise) and the constraints on their parameters. Swish is a single and continuous function and the parameter $\gamma$ has no constraints. ULU and AULU are piecewise functions and the \textbf{parameters must be positive}. The learnable parameter $\beta$ could be negative during training progress, so taking its square forces it to be positive. The default parameters settings in ULU below are based on ~\cref{eq:ULU}.

\begin{figure}[t!]
    \centering
    \begin{subfigure}[b]{\textwidth}
        \centering
        \includegraphics[width=0.3\linewidth]{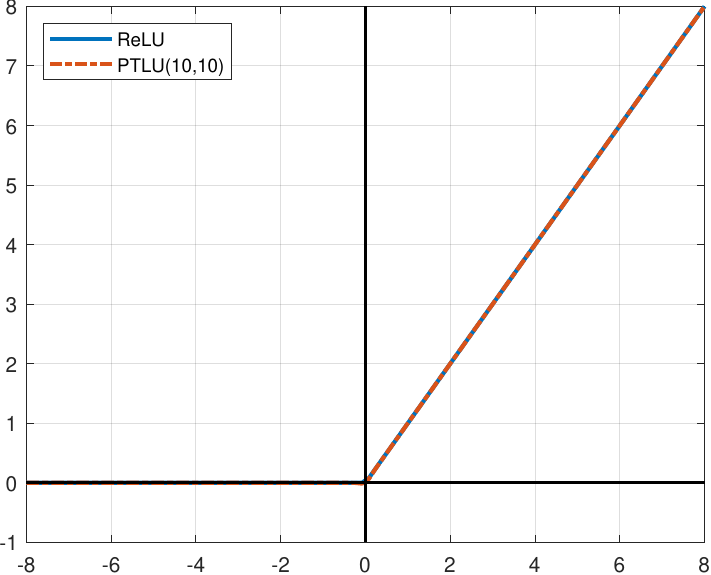}
        \hfill
        \includegraphics[width=0.3\linewidth]{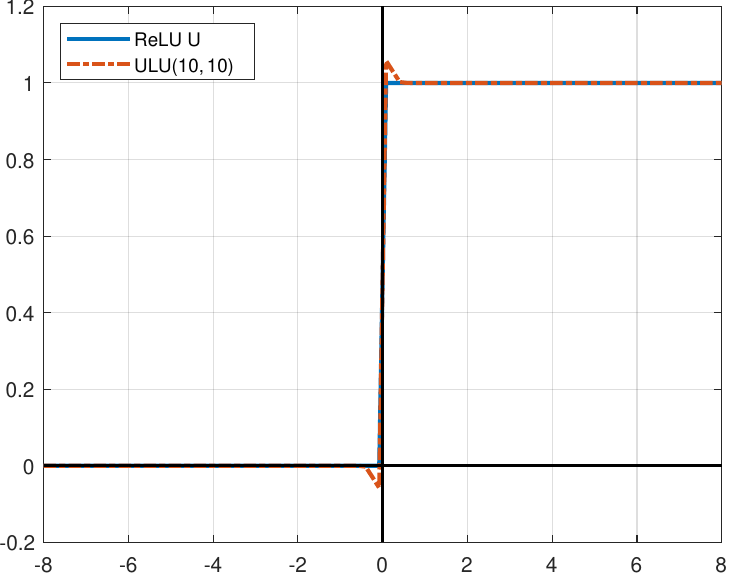}
        \hfill
        \includegraphics[width=0.3\linewidth]{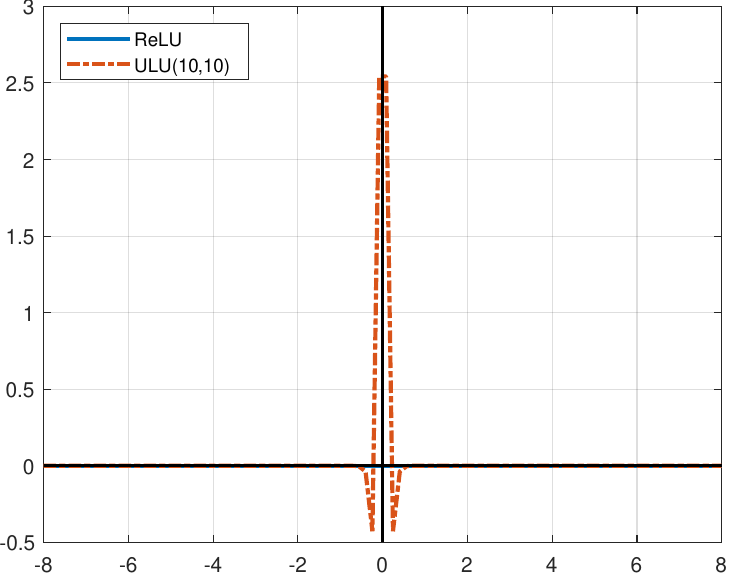}
        \caption{ReLU $\approx$ ULU(10,10)}
    \end{subfigure}
    

    \begin{subfigure}[b]{\textwidth}
        \centering
        \includegraphics[width=0.3\linewidth]{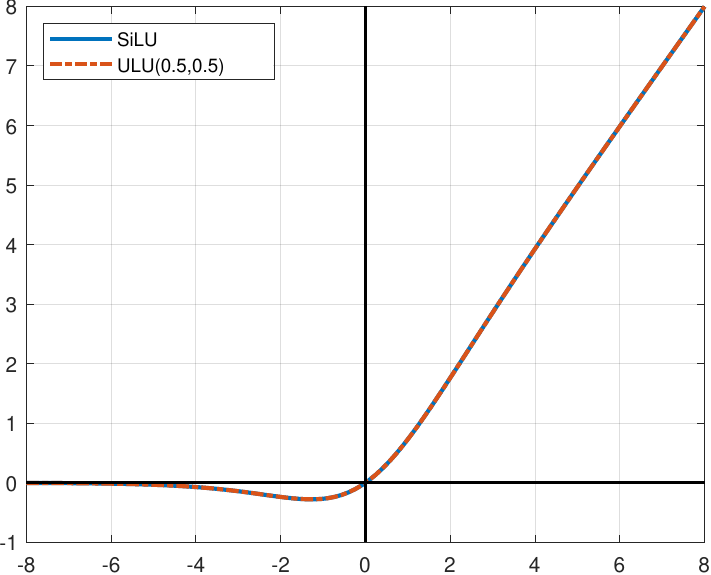}
        \hfill
        \includegraphics[width=0.3\linewidth]{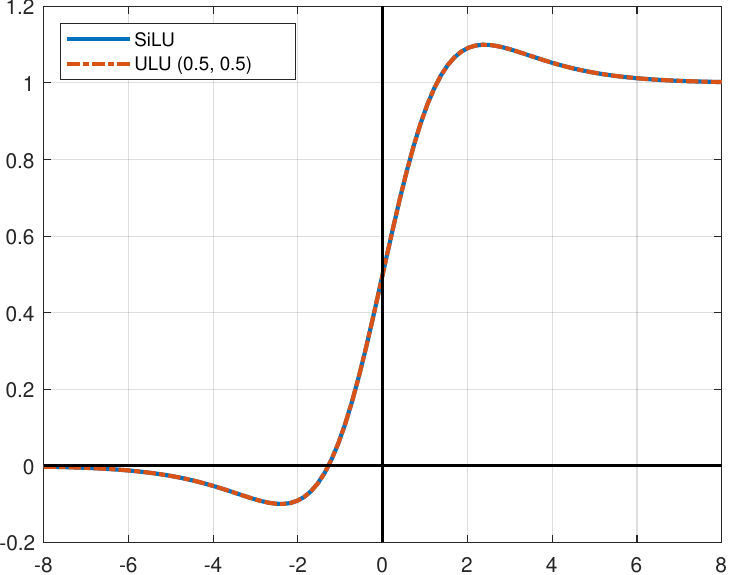}
        \hfill
        \includegraphics[width=0.3\linewidth]{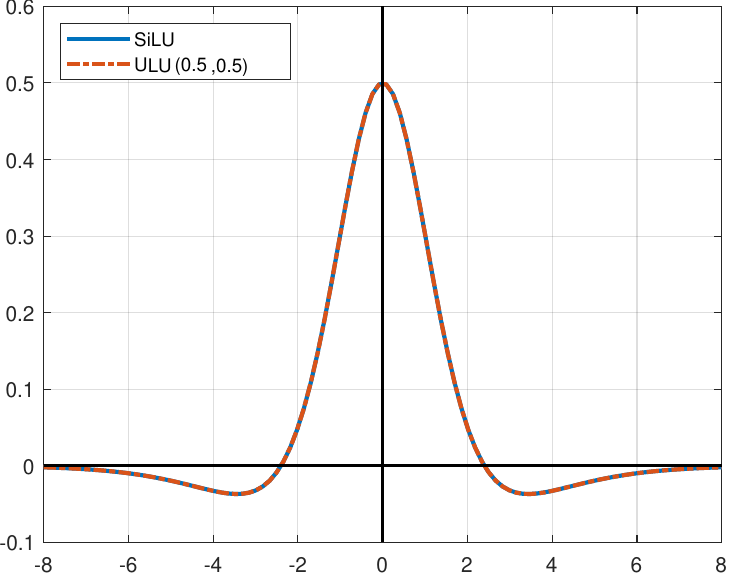}
        \caption{SiLU $=$ ULU(0.5,0.5)}

    \end{subfigure}
        
    \begin{subfigure}[b]{\textwidth}
        \centering
        \includegraphics[width=0.3\linewidth]{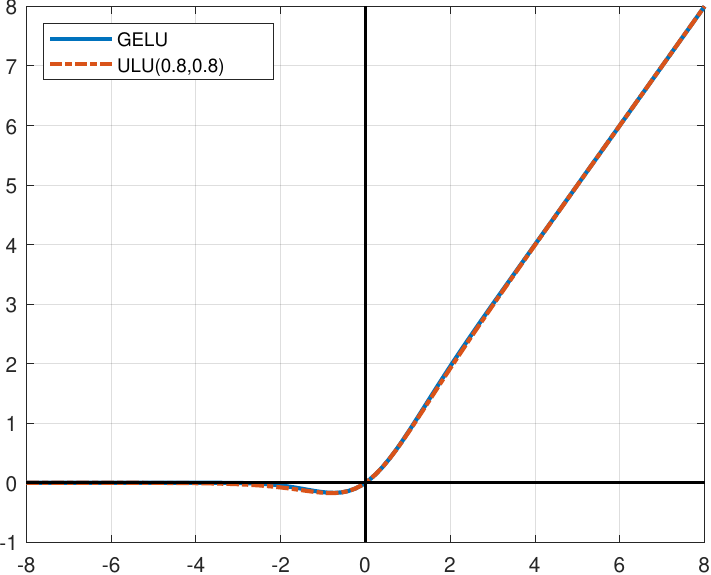}
        \hfill
        \includegraphics[width=0.3\linewidth]{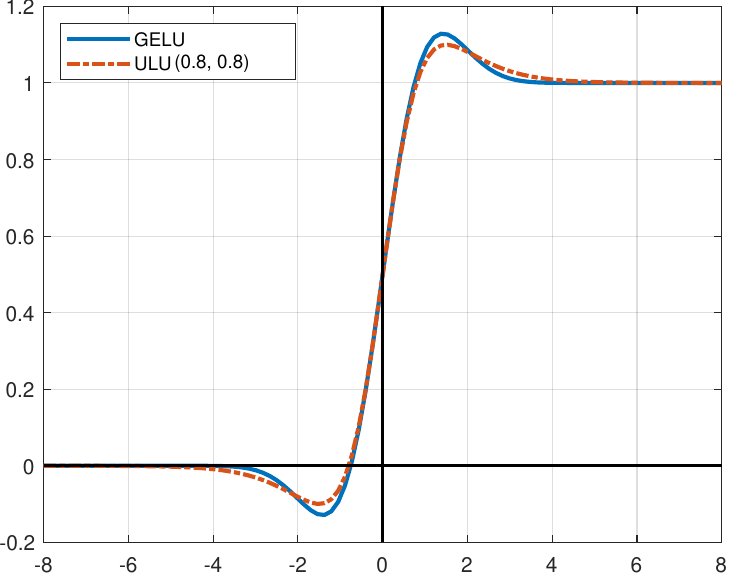}
        \hfill
        \includegraphics[width=0.3\linewidth]{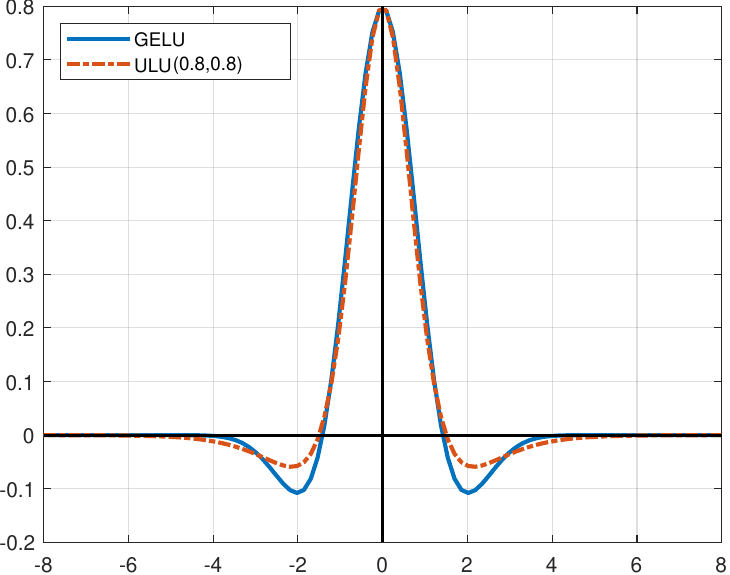}
        \caption{GELU $\approx$ ULU(0.8,0.8)}
    \end{subfigure}
    

    \begin{subfigure}[b]{\textwidth}
        \centering
        \includegraphics[width=0.3\linewidth]{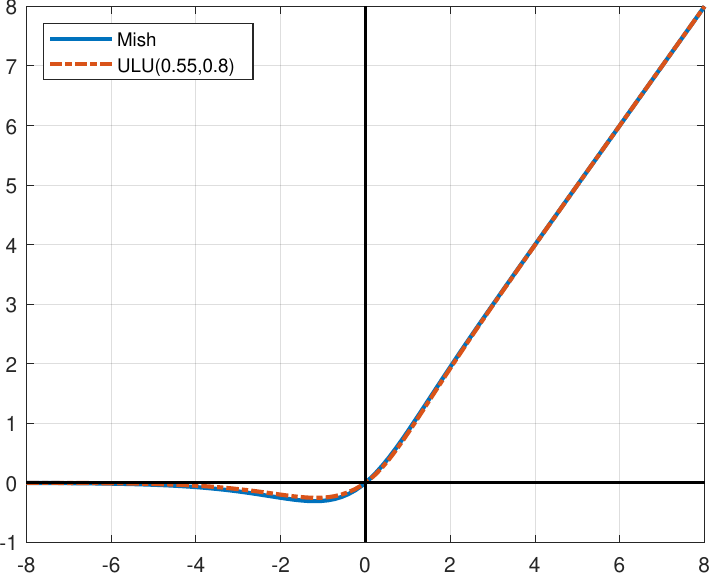}
        \hfill
        \includegraphics[width=0.3\linewidth]{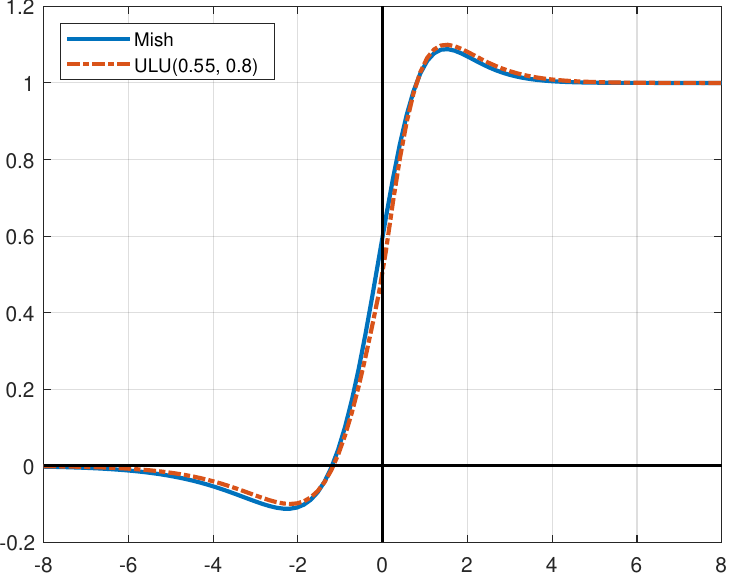}
        \hfill
        \includegraphics[width=0.3\linewidth]{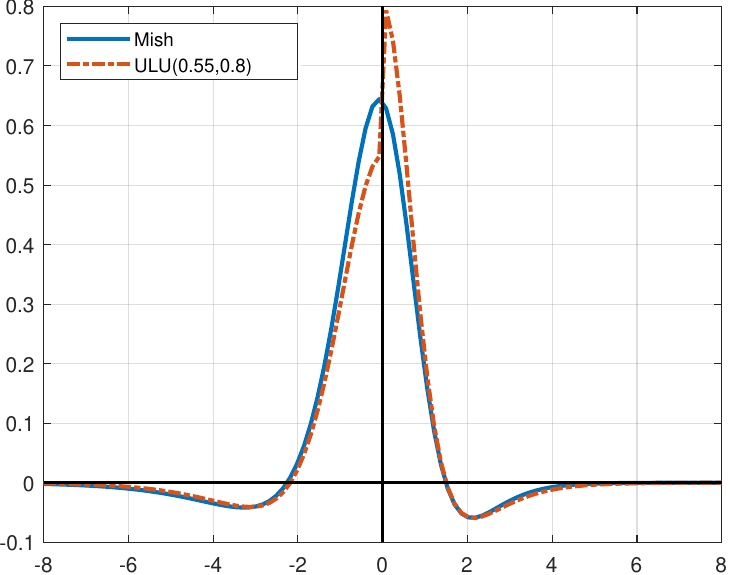}
        \caption{Mish $\approx$ ULU(0.55,0.8)}

    \end{subfigure}

    \caption{Common activation functions (Left) compared to ULU with different $(\alpha_1, \alpha_2)$ and their $1^{st}$ (Middle) and $2^{nd}$ (Right) derivatives.}
\label{fig:acts}
\end{figure}

\subsection{Properties}
\begin{figure}[t]
\centering
\includegraphics[width=0.8\textwidth]{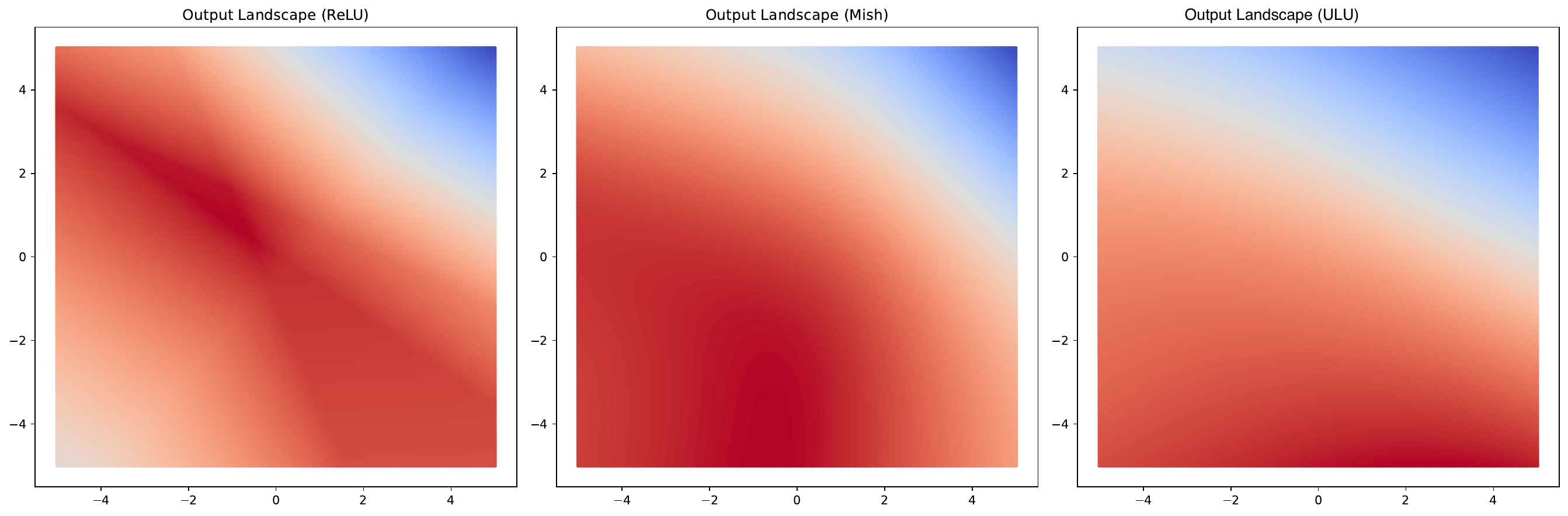} 
\caption{Comparison of the output landscapes of ReLU (Left), Mish (Middle) and ULU (Right) activation functions}
\label{fig:losslanscape}
\end{figure}

ULU and AULU are unbounded above yet bounded below. It is smooth, non-monotonic and differentiable. ULU and AULU also retain a small portion of negative weights. The advantageous properties are as follows:

\begin{itemize}
    \item \textbf{Differentiability:} Avoids singularities and issues during gradient-based optimization. Unlike non-differentiable ReLU. The first derivatives of the smooth function AULU is continuous. 

    \item \textbf{Unbounded Above}: Avoids saturation during training which occurs with bounded functions like \textit{sigmoid} and \textit{tanh} having near-zero gradients. Being unbounded above like ReLU, ULU and AULU mitigate this issue. Its positive portion behaves approximately linearly (Figure \cref{fig:acts}), making it suitable as an activation.

    \item \textbf{Bounded Below:} Provides strong regularization effects. However, ReLU dies when receiving negative inputs. By preserving some negative values, ULU and AULU reduce this problem and improves performance and gradient flow. 
    
    \item \textbf{Smooth:} Smooth loss landscapes enable easier training and generalization. Figure \cref{fig:losslanscape} shows output landscapes of a 6-layer network with ReLU, Mish and AULU activation functions. The output topology reflects the loss shape. By inputting grid points $(x,y)$ and plotting network outputs, we visualize the output landscape. ReLU exhibits jagged, sharp transitions versus the smoother contours of AULU. This empirically demonstrates how AULU facilitates optimized training, aligning with its performance gains over ReLU.

   \item \textbf{Diversity:} By varying the value of $(\alpha_1, \alpha_2)$, the activation function can exhibit different response behaviors in the negative and positive regions, as shown in Figures \cref{fig:acts}. ULU(0.8, 0.8) exhibits properties close to GELU.  ULU(0.5, 0.5) equals SiLU. It can be naturally inferred that ULU(0.5, 0.8) approximates SiLU and GELU in the positive and negative regions, respectively. ULU(0.55, 0.8) behaves similarly to Mish. It's may helpful to explain Mish's strong performance, as Mish$\approx$ULU(0.55, 0.8) which means Mish has different patterns in the positive versus negative domains. The tunable hyper-parameters of ULU can be flexibly set to mimic different activation functions in separate regions to suit different task and model requirements.
   

\end{itemize}

\section{Experiments}

In this section, we benchmark our proposed ULU activation against state-of-the-art architectures across diverse tasks. Additionally, ablation studies on MNIST and CIFAR-10 provide further insights about ULU. Overall, ULU achieved superior performance over existing activation functions for most tasks, indicating its versatility.

\subsection{Tuning ULU($\alpha_1, \alpha_2$)}
\begin{figure*}[t]
    \centering
    \includegraphics[width = 0.45\textwidth]{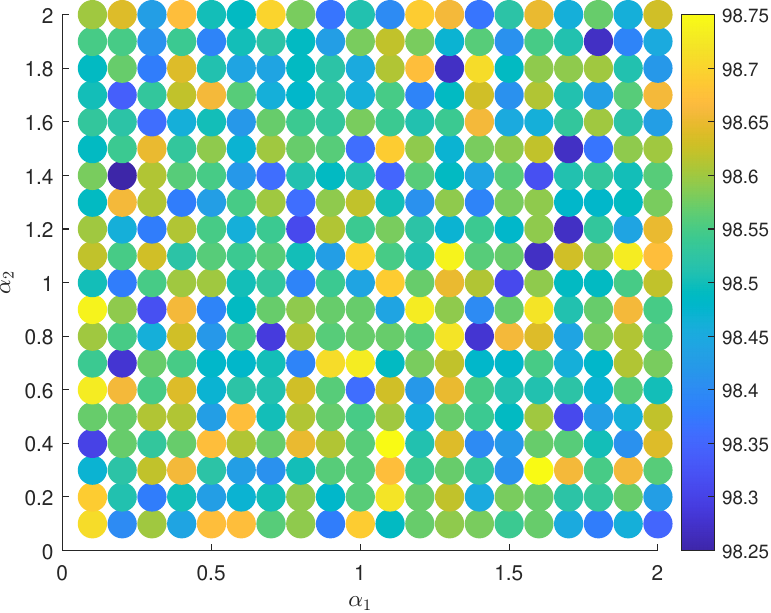}
    \hspace{0cm}
    \includegraphics[width = 0.45\textwidth]{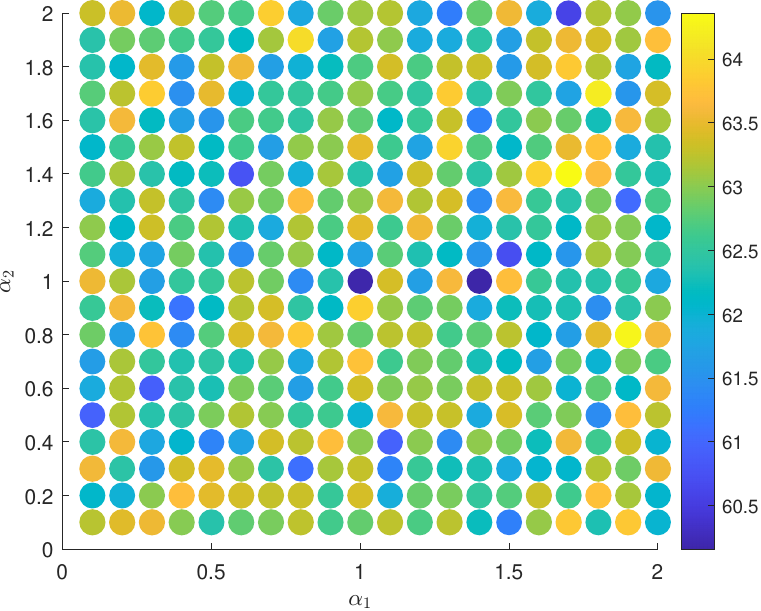}
    \caption{Accuracy of ULU($\alpha_1,\alpha_2$) with varying $\alpha_1,\alpha_2$ combinations on MNIST (Left) and CIFAR-10 (Right) datasets. The depth of color shading represents relative accuracy levels. }
      \label{fig:a1a2}
\end{figure*}

In order to determine the optimal combination of $\alpha_1$ and $\alpha_2$, we conducted experiments on MNIST \cite{LeCun1998GradientbasedLA} and CIFAR-10 \cite{krizhevsky2009learning} using a simple convolution network for image classification. The values of $\alpha_1$ and $\alpha_2$ ranged from 0.1 to 2.0. The classification accuracy results for different $(\alpha_1, \alpha_2)$ pairs are visualized as circles with varying color shades in the Figure \ref{fig:a1a2}. However, analysis of the results did not reveal clear patterns or correlations between the tunable parameters and accuracy. While certain $(\alpha_1, \alpha_2)$ pairs achieved high performance, the global optimum was difficult to deduce from the experiments.  More experiments are necessary to systematically analyze the impact of $\alpha_1$ and $\alpha_2$ on ULU's performance as well as characterize the settings that maximize its effectiveness for different models and tasks in the future.

\subsection{Statistical Analysis}

To further analyze the statistical significance of ULU's superior image classification accuracy, controlled experiments were conducted on CIFAR-10 \cite{krizhevsky2009learning} using ResNet-18 \cite{he2016deep} without pre-trained weights with the SGD \cite{robbins1951stochastic} optimizer and changing only the activation function. The SGD optimizer was configured with a momentum of $0.9$, weight decay of $5\times 10^-5$, and learning rate scheduler using a warming-up policy \cite{he2019bag}. 

As summarized in Table \ref{tab:sota_cifar10}, ULU(0.3,0.8) achieved the highest mean accuracy  of 88.7\% over 10 runs, surpassing Swish and Mish. ULU also attained the lowest accuracy standard deviation of 0.321, exhibiting consistency despite varying conditions. Moreover, ULU slightly exceeded top-performing GELU, empirically validating its optimization advantages. ULU's adaptive formulation enables mimicking Swish and GELU's benefits. 

In summary, comparative studies statistically verify ULU's effectiveness and consistency, underscoring its potential as a versatile activation for computer vision tasks .

\begin{table}[htp]\footnotesize
    \centering
    \renewcommand{\arraystretch}{1.3}
    \begin{tabular}{cccc}

    \toprule
    Activation & $\mu_{acc}$  & $\sigma_{acc}$ \\
    \midrule
    \rowcolor{LightCyan}
           ULU(0.3,0.8)  & \textbf{88.7$\%$} &  0.321 \\[.5ex]
            
 	    Mish \cite{Misra2019MishAS}  & 87.9$\%$  & 0.332 \\[.5ex]
			Swish \cite{Ramachandran2017SearchingFA}  & 88.0$\%$  & 0.330 \\[.5ex]
			GELU \cite{Hendrycks2016GaussianELU}  & 88.3$\%$  & 0.356 \\[.5ex]
			ReLU \cite{Nair2010RectifiedLU} & 86.7$\%$ &  0.384 \\[.5ex]
			ELU \cite{Clevert2015FastAA}  & 84.6$\%$ &  0.416 \\[.5ex]
			Leaky ReLU \cite{Maas2013RectifierNI}  & 87.1$\%$ &  0.347  \\[.5ex]
			SELU \cite{Klambauer2017SelfNormalizingNN}  & 81.7$\%$ &  0.452 \\[.5ex]
			RReLU \cite{xu2015empirical}  & 86.1$\%$  & 0.443\\[.5ex]
    
    \bottomrule
    \end{tabular}\\
    \caption{Top-1\% Accuracy values of different activation functions on image classification of CIFAR-10 dataset using a Resnet-18 for 10 runs.}
    \label{tab:sota_cifar10}

\end{table}

\subsection{Image Classification}

\paragraph{ULU's two parameters set to constants.}

Table \ref{tab:cifar10top1} summarizes the top-1\% accuracy on CIFAR-10 image classification across 9 leading convolutional neural network architectures. Controlled experiments isolated the impact of replacing the native activation with ULU, Mish or ReLU.

Examining the results vertically for each model, ULU consistently achieves the highest accuracy, outperforming both Mish and ReLU. For example, in DenseNet-121, ULU reaches 80.4\% accuracy, exceeding ReLU by 6.6\% and surpassing Mish by 1.9\%.  In Mobilenet-v2, ULU also improves accuracy over ReLU by up to 6.6\% and outperformed Mish by 0.8\%, demonstrating optimization advantages over the state-of-the-art activation.

Analyzed horizontally, the margins between ULU and ReLU are noteworthy, ranging from 1.6\% in EfficientNet-B0 to up to 9.1\% in ShuffleNet-v2. This demonstrates ULU's broad efficacy across diverse model complexities. The gaps between ULU and Mish are smaller but still substantial, especially for compact models like ShuffleNet where more challenging optimization exacerbates activation limitations. Some training progress are shown in Figures \ref{fig:curves}.

In conclusion, these controlled experiments validate ULU as an advantageous activation for computer vision tasks. The results strongly demonstrate ULU's effectiveness in boosting performance given the same model capacity, making it an attractive plug-and-play replacement for existing static activations.

\begin{table}[!t]\footnotesize
    \centering
    \renewcommand{\arraystretch}{1.3}
    \begin{tabular}{ccca}

    \toprule
    \textbf{Methods} & \textbf{ReLU} & \textbf{Mish} & \textbf{ULU (0.3,0.8)}\\[.5ex]
    \midrule
    DarkNet-19\cite{redmon2016yolo} & 85.8 & 87.5 & \textbf{88.2}\\[.5ex]
    
    Resnet-34\cite{he2016deep} & 82.8 & 84.2 & \textbf{85.4}\\[.5ex]
    
    WideResnet-50-2\cite{zagoruyko2016wide} & 72.2 & 81.9 & \textbf{82.6}\\[.5ex]
    
    ShuffleNet-v2\cite{ma2018shufflenet} & 70.4 & 76.6 & \textbf{79.5}\\[.5ex]
    
    Inception-v3\cite{szegedy2017inception}& 69.8 & 74.3 & \textbf{75.1}\\[.5ex]
    
    DenseNet-121\cite{huang2017densely} &  73.8 & 78.5 & \textbf{80.4}\\[.5ex]
     
    MobileNet-v2\cite{sandler2018mobilenetv2}&  77.1 & 82.9 & \textbf{83.7}\\[.5ex]

    SqueezeNet\cite{iandola2016squeezenet} &  60.6 & 65.3 & \textbf{65.6}\\[.5ex]

    EfficientNet-B0\cite{tan2019efficientnet} &  66.0 & 67.4 & \textbf{67.6}\\[.5ex]
    
    \bottomrule
    \end{tabular}\\
    \caption{Top-1\% accuracy comparison between ULU, Mish, and ReLU based on image classification of CIFAR-10 across various models.}
    \label{tab:cifar10top1}

\end{table}

\begin{table}[!t]\footnotesize
    \centering
    \renewcommand{\arraystretch}{1.3}
    \begin{tabular}{ccca}

    \toprule
    \textbf{Methods} & \textbf{ReLU} & \textbf{Mish} & \textbf{ULU (0.3,0.8)}\\[.5ex]
    \midrule
    DarkNet-19\cite{redmon2016yolo} & 37.7 & 46.1 & \textbf{48.5}\\[.5ex]
    
    Resnet-34\cite{he2016deep} & 46.2 & 49.0 & \textbf{50.0}\\[.5ex]
     
    MobileNet-v2\cite{sandler2018mobilenetv2}&  37.3 & 44.6 & \textbf{45.7}\\[.5ex]

    \bottomrule
    \end{tabular}\\
    \caption{Top-1\% accuracy comparison between ULU, Mish, and ReLU  based on image classification of CIFAR-100 across various models.}
    \label{tab:cifar100top1}

\end{table}


    
    

\begin{figure}[t!]
    \centering
    \includegraphics[width = 0.23\textwidth,height = 0.23\textwidth]{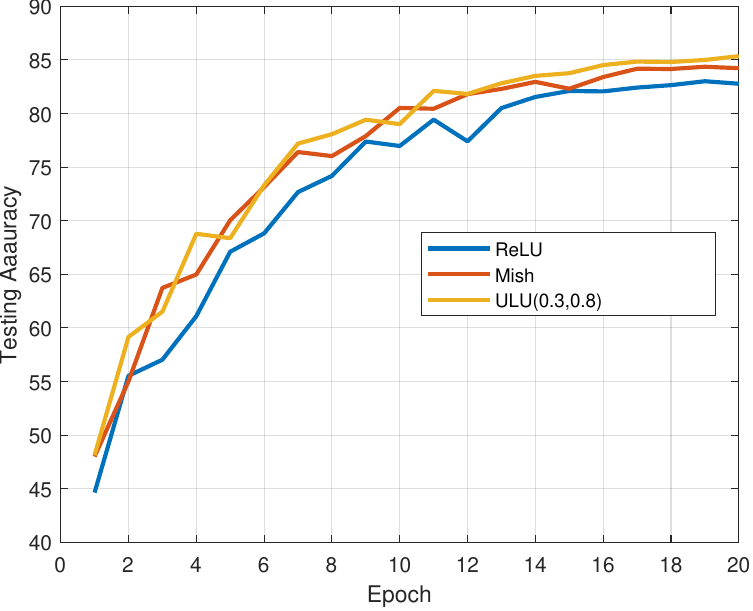}
    \hspace{0cm}
    \includegraphics[width = 0.23\textwidth,height = 0.23\textwidth]{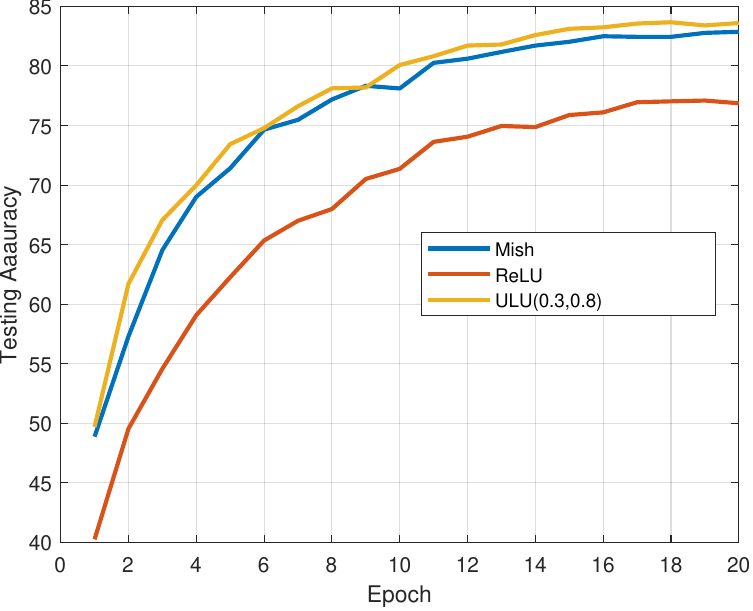}
    \hspace{0cm}
    \includegraphics[width = 0.23\textwidth,height = 0.23\textwidth]{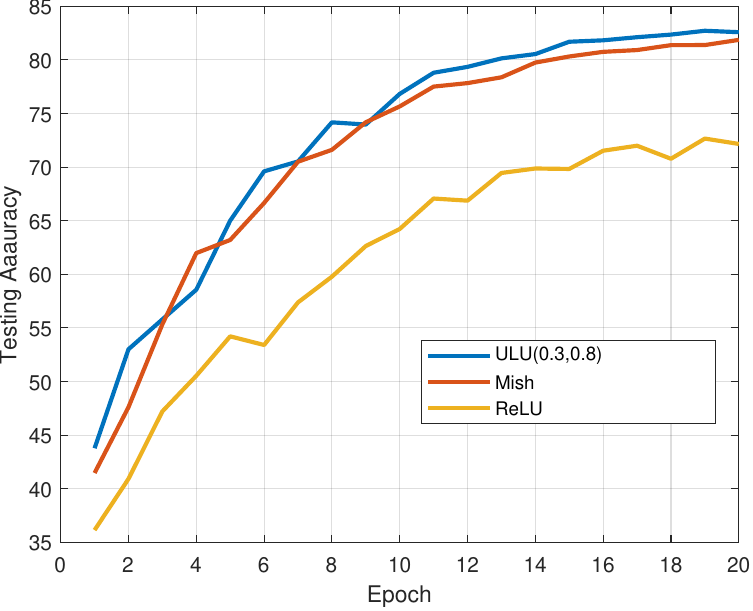}
    \hspace{0cm}
    \includegraphics[width = 0.23\textwidth,height = 0.23\textwidth]{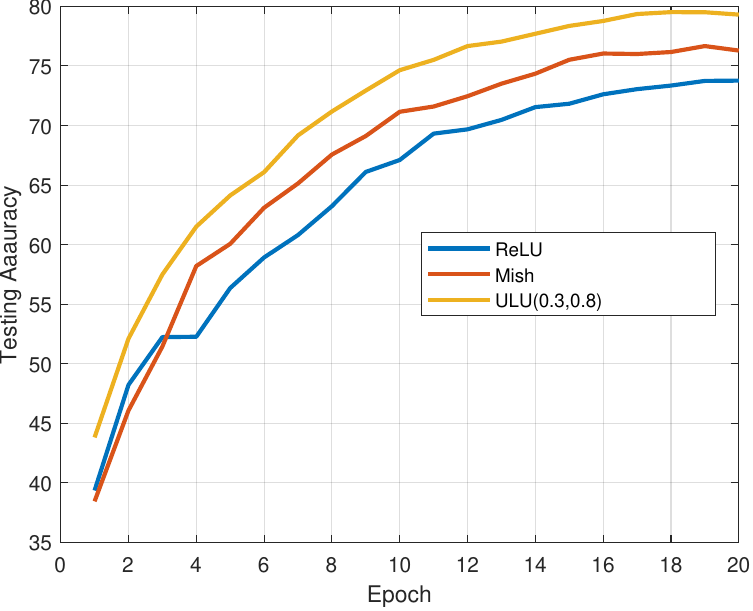}
    \caption{Training curves of ResNet34 (Left), MobileNet, Wide-Resnet50, Shufflenet (Right) with activation functions: ReLU, Mish and ULU}%
    \label{fig:curves}
\end{figure}

\paragraph{AULU's two parameters set to be learnable}

We set AULU's two parameters to be learnable. In the training process, we monitored and recorded the evolution of two hyperparameters' values ($\beta_1, \beta_2$) of AULU.  We observe that the squared values of the two hyperparameters exhibit a significant discrepancy for the pure CNN model. In contrast, the difference of squared values for the pure Transformer model are closely to 0. 


\begin{figure}[!t]
    \centering
     \includegraphics[width = 0.5\textwidth,height = 0.5\textwidth]{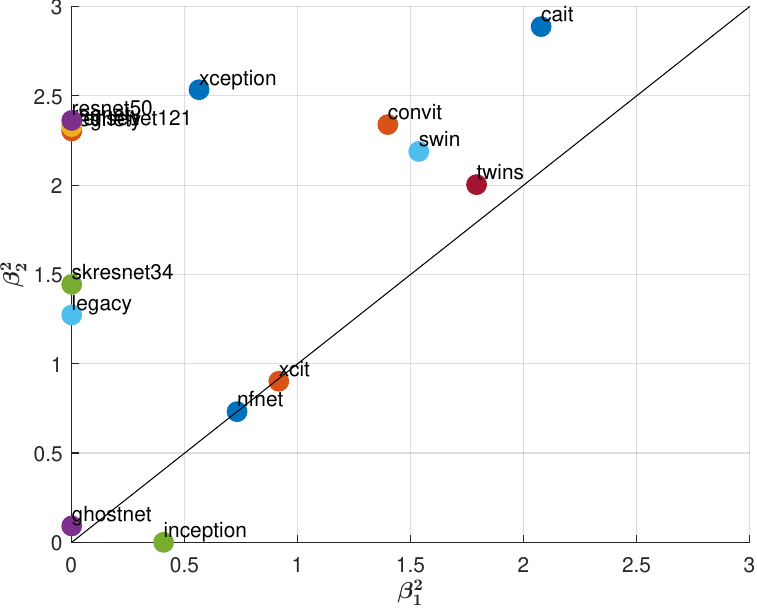}
\caption{The final two  parameters $\beta_1^2, \beta_2^2$ in AULU for different models.}
    \label{fig:beta12}
\end{figure}

In the figure \ref{fig:beta12}, we can observe that models based on Transformers are clustered closer to the $y=x$ line, while models based on CNNs are more scattered and farther away from the $y=x$ line. Therefore, when defining the LIB metric as the absolute difference between the two parameters, CNN models exhibit larger LIB values, aligning with the empirical observation that CNN models possess high inductive biases. Conversely, Transformer models have relatively small LIB values, corroborating the empirical evidence that Transformers lack significant inductive biases. The LIB value reflects the degree to which a model differentially treats the positive and negative regions. This quantitative characterization of inductive biases through LIB is a significant contribution of AULU.

We define a \textbf{Like Inductive Bias (LIB)} metric to quantitatively measure the inductive bias of the model, formulated as follows:. 
\begin{equation}
    LIB = |\beta_1^2 - \beta_2^2|
    \label{eq:lib}
\end{equation}

\subsection{Object Detection}

Object detection experiments evaluated ULU against native Leaky ReLU in YOLOv3 models \cite{redmon2018yolov3}  on Pascal VOC2012 \cite{everingham2010pascal}. Controlled tests only substituted the activation. ULU consistently achieved higher Mean Average Precision (MAP), improving YOLOv3 MAP@0.5 by 5\% and MAP@0.5:0.95 by 8.3\%. Similar gains occurred for tiny YOLOv3, showing broad effectiveness. Again, ULU validates superiority as a drop-in replacement, underscoring the activation's importance. Tunable ULU enables optimization for complex detection. In summary, ULU demonstrates potential to enhance modern detectors over ReLU-based activation functions. Adoption can yield noticeable accuracy improvements given fixed architectures.

\begin{table}[ht]\footnotesize
    \centering
    \renewcommand{\arraystretch}{1.3}
    \begin{tabular}{cccc}

    \toprule
    \textbf{Model} & \textbf{Activations} & \textbf{MAP@.5} & \textbf{MAP@.5:.95}\\[.5ex]
    \midrule
    
    YOLOv3 & LeakyReLU & 72.2 & 44.3\\[.5ex]
    
    \rowcolor{LightCyan}
    
    YOLOv3 & ULU(0.5,0.8) & \textbf{77.1} & \textbf{52.6}\\[.5ex]
     
    \midrule
    
    YOLOv3 Tiny & LeakyReLU & 49.6 & 20.8\\[.5ex]
    
    \rowcolor{LightCyan}
    
    YOLOv3 Tiny & ULU(0.5,0.8) & \textbf{52.4} & \textbf{21.7}\\[.5ex]

    \bottomrule
    \end{tabular}\\
    \caption{Mean Average Precision scores for Leaky ReLU and ULU in YOLOv3 models on the Pascal VOC2012 dataset.}
    \label{tab:detection}
\end{table}

\section{Conclusion}

In this work, we have introduced the Unified Linear Unit (ULU), a novel paradigm that unifies a broad spectrum of common activation functions. We have demonstrated that prominent activations, including GELU, ReLU, and Mish, can be subsumed as special cases within the ULU framework, each corresponding to a specific configuration of its two core hyperparameters. The defining characteristic of ULU is its inherent structural asymmetry, which facilitates distinct computational responses to positive and negative inputs, offering a more flexible and expressive architectural component. 

We proposed the Adaptive Unified Linear Unit (AULU), which extends the ULU concept by rendering its parameters learnable. which enables a novel analytical approach. We introduce the Like Inductive Bias (LIB) index, quantitatively defined as the absolute difference between the two learned parameters of the AULU. We think LIB could potentially serve as a novel diagnostic signature of a model's internal state and provide a quantitative measure of the model's alignment and safety.

%
%
\bibliographystyle{splncs04}
\bibliography{main}
\end{document}